\documentclass{ecai} 
\usepackage{latexsym}
\usepackage{amssymb}
\usepackage{amsmath}
\usepackage{amsthm}
\usepackage{booktabs}
\usepackage{enumitem}
\usepackage{graphicx}
\usepackage{color}
\usepackage{tabularx}
\usepackage[table]{xcolor}
\usepackage{stfloats}
\usepackage{multicol}

\newcommand{\BibTeX}{B\kern-.05em{\sc i\kern-.025em b}\kern-.08em\TeX}
\newcolumntype{Y}{>{\centering\arraybackslash}X}
\definecolor{green}{HTML}{c5e0b3}

\begin{document}

%%%%%%%%%%%%%%%%%%%%%%%%%%%%%%%%%%%%%%%%%%%%%%%%%%%%%%%%%%%%%%%%%%%%%%%%

\begin{frontmatter}

%%% Use this command to specify your submission number.
%%% In doubleblind mode, it will be printed on the first page.

\paperid{3} 

%%% Use this command to specify the title of your paper.

% \title{Resilient Movement Planning for Continuum Robots}

\title{Resilient Movement Planning for Continuum Robots \\[12pt] 
\small \normalfont (The author's version was accepted by Multi-Objective Decision Making Workshop at ECAI 2024 \\ the final publication is available at \url{https://modem2024.vub.ac.be/papers/MODeM2024_paper_3.pdf})}

%%% Use this combinations of commands to specify all authors of your 
%%% paper. Use \fnms{} and \snm{} to indicate everyone's first names 
%%% and surname. This will help the publisher with indexing the 
%%% proceedings. Please use a reasonable approximation in case your 
%%% name does not neatly split into "first names" and "surname".
%%% Specifying your ORCID digital identifier is optional. 
%%% Use the \thanks{} command to indicate one or more corresponding 
%%% authors and their email address(es). If so desired, you can specify
%%% author contributions using the \footnote{} command.

\author[A]{\fnms{Oxana}~\snm{Shamilyan}\orcid{0000-0002-0961-9059}\thanks{Corresponding Author. Email: shamilyan@ihp-microelectronics.com}}
\author[A]{\fnms{Ievgen}~\snm{Kabin}}
\author[A,B]{\fnms{Zoya}~\snm{Dyka}}
\author[A,B]{\fnms{Peter}~\snm{Langendoerfer}} 

\address[A]{IHP – Leibniz-Institut für innovative Mikroelektronik}
\address[B]{BTU Cottbus-Senftenberg}

%%% Use this environment to include an abstract of your paper.

\begin{abstract}
This paper presents an experimental study of resilient path planning for continuum robots taking into account the challenge of multi-objective optimisation. To do this, we used two well-known algorithms for path planning, namely Genetic algorithm and A* algorithm, and modified them by adding the Analytical Hierarchy Process algorithm for paths’ evaluation. In our experiment the Analytical Hierarchy Process considers four different criteria, i.e. distance, motors damage, mechanical damage of the robot’s arm and accuracy, each considered to contribute to the resilience of a continuum robot. The use of different criteria is necessary to increasing the time to maintenance operations of the robot. The experiment provides us two insights. First, the improved algorithms with multi criteria path-planning show better performance than their classical versions. Second, when comparing improved GA and A* algorithm, A* shows superior performance. 
\end{abstract}

\end{frontmatter}

%%%%%%%%%%%%%%%%%%%%%%%%%%%%%%%%%%%%%%%%%%%%%%%%%%%%%%%%%%%%%%%%%%%%%%%%

\section{Introduction}

Continuum robots are a new type of robots providing a high degree of flexibility when it comes to reaching tricky positions, as they are not limited in their movements by joints. Instead, they have infinite degrees of freedom. On the one hand, this makes them ideal candidates for application areas such as inspection tasks, handling of unstructured objects, and minimally invasive surgery \cite{shamilyan_meco_2022, shamilyan_access_2023}. On the other hand, the high flexibility of these robots presents a challenge in terms of controlling their movements. \cite{shamilyan_access_2023} provides a comprehensive survey of the most popular continuum robots types, motion control models and examples of research groups that are focused particularly on the study of motion control problems of the continuum robots. One of the most popular motion control model is the Cosserat theory \cite{cosserat}. Many researchers \cite{intro_3, intro_1, intro_2, intro_5, intro_4}  utilize it for modelling motion control, which considers different kinds of deformations of physical objects, but requires a long time and significant computation costs. To reduce computation costs, \cite{intro_2}  used a discrete Cosserat approach, means they processed the continuum robot prototype as a finite set of constant pieces, making the solution discrete and easy to process. \cite{intro_6}  provides a comparison of different motion models and demonstrates that the Cosserat theory can be used for the tasks where the high accuracy is needed, but is not suitable for real-time applications.

Path planning is essential for the continuum robots, especially when deployed as autonomous systems in unknown environments. Under these conditions the path-planning should take into account not only the start and the end point, but also the state of the environment and the state of the system itself. Such awareness helps the system to make a decision and generate the optimal path. Here we mean by optimal not essentially the shortest path but the one with the least negative impact on the continuum robot. In order to enable a robot to properly assess more conditions than just the path length, we modified path planning algorithms and added an Analytical Hierarchy Process (AHP) decision-making algorithm. The latter generates the weights to evaluate the generated path, to find the optimal solution. In our experiments we used Genetic algorithm and A* algorithm. 

The rest of this paper is structured as follows. In section II we provide information about the methods we used, namely AHP, Genetic algorithm and A* algorithm. Section III describes our experimental setup. In section IV we provide information about our experiments and results. This paper ends with a short conclusion.

\section{Methods}
This section describes the methods used in our experiment, which aims at comparing two path planning algorithms: Genetic algorithm and A* algorithm, when more criteria than the path length have to be considered. We made changes to the classical representations of both algorithms to suit our needs, primarily by adding multiple criteria for assessing the generated path. The criteria weights were generated using the AHP decision-making method.

\subsection{Analytical Hierarchy Process}
AHP is the most common multi-criteria decision-making algorithm developed by T.L. Saaty in 1980 \cite{saaty_analytic_1980}. It makes decisions based on pairwise comparison of alternatives or criteria depending on their relative importance levels (Table \ref{tab:Table1}). AHP is applied in various fields such as business, marketing, risk analysis for safety enhancement, environmental location selection, performance evaluation, and path planning. 

A full description of the algorithm can be found in \cite{ga_basics}. Below we briefly describe each step:

\begin{enumerate}
  \item Problem analysis;
  \item Generation of the hierarchy structure; 
  \item Definition of the relative importance of the objectives (Table \ref{tab:Table1});
  \item Weight importance calculation; 
  \item Evaluation of the consistency indexes for objectives; 
  \item Evaluation of alternatives with reference to each objective.
\end{enumerate}

For the analysis, we established four criteria that are crucial for enhancing the resilience of a continuum robot. This includes reducing or avoiding mechanical damage and increasing the time to the next maintenance operation. Table \ref{tab:Table2} provides information about all four criteria, including its name, description of what each criterion estimates, and description of resilience properties that are reasoning the choice of criteria.

\begin{table}[ht]
  \centering
  \caption{The pairwise comparison scale}
  \label{tab:Table1}
  \begin{tabular}{|c|c|}
    \hline
    \textbf{Importance} & \textbf{Definition} \\
    \hline
    1 & Equal importance of elements \\
    \hline
    3 & Moderate importance of one element over another \\
    \hline
    5 & Strong importance of one element over another \\
    \hline
    7 & Very strong importance of one element over another \\
    \hline
    9 & Extreme importance of one element over another \\
    \hline
    2, 4, 6, 8 & Intermediate values between two adjacent judgment \\
    \hline
  \end{tabular}
\end{table}

\begin{table}[t]
  \centering
  \caption{Defined criteria}
    \begin{tabularx}{\columnwidth}{|p{0.2cm}|X|X|X|}
    \hline
    \textbf{\#} & \textbf{Criteria name} & \textbf{Description} & \textbf{Resilience properties} \\
    \hline
    1 & distance & estimates length of the generated path & energy and time savings\\
    \hline
    2 & motor damage & estimates damage of setup’s motors & extended time to maintenance of the motors\\
    \hline
    3 & {mechanical damage} & {estimates damage will be caused to robot tendons after the robot follows the generated path} & {extended time to maintenance of robot wire cables}\\
    \hline
    4 & robot’s tip accuracy & estimates how accurately the robot’s tip is at the goal point & accurate work\\
    \hline
  \end{tabularx}
  \label{tab:Table2}
\end{table}

\subsection{Genetic Algorithm}
The genetic algorithm (GA) is a global search and optimisation technique. It was first introduced by John H. Holland in \cite{ga_basics}. The algorithm is inspired by Darwin’s theory and works on the principle of natural selection. GA can be applied to a wide range of disciplines, such as mathematics, medicine \cite{ghaheri_applications_2015}, and engineering \cite{bhoskar_genetic_2015}.

The algorithm includes five main steps: initialisation and genetic operators – evaluation, selection, crossover, and mutation. GA continues to repeat genetic operators, creating new generations, until the terminal condition is met. For our experiment we chose population size 50 and generation number 3 as optimal values to conduct the experiment.\footnote{We run a few experiments to determine reasonable parameters which we omit here for better readability. The number reported provided the best result.} Further we provide details about each of the steps of the algorithm and specify which techniques we used for the experimental part.

\subsubsection{Initialisation of Population}
The population has a fixed size and consists of \textit{chromosomes}, each of which consists of \textit{genes}. The chromosome length is not necessarily fixed. Each chromosome represents a possible solution to the problem. During the initialisation step, genes are generated and combined into chromosomes until the whole population is ready. 

In our implementation we used an improved random technique to initialize the population. A database of possible connections between nodes was created and used for the initialisation. A starting node is always defined. Each next node \textit{n}$_\textit{i}$ is randomly chosen among the possible connections of the previous node \textit{n}$_\textit{i-1}$.

\subsubsection{Evaluation}
Each chromosome in the population is evaluated using the \textit{fitness function} that indicates how well each chromosome fits the current problem. A correctly composed fitness function increases the algorithm’s output quality. The fitness function can contain one assessment function or set of them.

For our experiment we use a \textit{multi-fitness function} which is calculated as a sum of the fitness functions for each criterion.

\begin{eqnarray}\label{eq:eq_1}
F = \sum_{i}w_if_i
\end{eqnarray}
where \textit{i} refers to the criterion from Table \ref{tab:Table2}, \textit{w}$_\textit{i}$ refers to the criterion weight obtained by the AHP algorithm and \textit{f}$_\textit{i}$ refers to fitness costs. 

\subsubsection{Selection}
During the selection step, the algorithm selects \textit{parent chromosomes} for \textit{crossover} and \textit{mutation}. The selection algorithm picks chromosomes to act as parents for the \textit{offspring population}. In our implementation of GA we use the Roulette Wheel Selection algorithm that is one of the most common and optimal solutions \cite{luca_roulette_2020}.

\subsubsection{Crossover}
The \textit{crossover} operator manipulates chromosomes in the population to create a diversity for future generations. In our implementation we use a single-point crossover, where two parent chromosomes are used to produce two offspring chromosomes. A separator (single-point) is randomly chosen to divide each parent into two parts. The first parts of both parents' chromosomes remain unchanged, while the second parts are swapped between each other to create the offspring.

\subsubsection{Mutation}
At the \textit{mutation} step, only a small part of the chromosome is changed. For our implementation we use a random mutation. A single gene of the parent chromosome is randomly changed to a randomly generated value. The parent chromosome to which the mutation operator is applied, also is randomly selected, so only a few chromosomes in the entire population are mutated. Whether or not a chromosome is mutated depends on the mutation rate value, which is set manually by the user before the algorithm begins to work. 

\subsection{Algorithm results}
After 3 generations (algorithm repetitions), the algorithm provides a final population that contains 50 paths. The path with the best fitness is considered as the resulting path.

\subsection{A* Algorithm}
The A* algorithm is a well-known path planning algorithm. It was proposed by Peter Hart, Nils Nilsson, and Bertram Raphael \cite{hart_formal_1968}. It is widely used in various domains, i.e. game development, logistics, and path planning for mobile robots \cite{barnouti_pathfinding_2016, guruji_time-efficient_2016, jianqin_research_2022, kim_development_2020,  zuo_hierarchical_2015}.

The core idea of the classical algorithm is to find the minimum distance between the starting and the goal nodes. A* achieves this by using heuristic search. It calculates the cost of moving from the current node n to its neighbour nodes, and creates a queue based on the cost results. The node with the lowest cost is given a higher position in the queue. The cost function is presented below:

\begin{eqnarray}\label{eq:eq_2}
f(n) = g(n) + h(n)
\end{eqnarray}
where g(n) denotes the cost from the starting node to the current node n and h(n) denotes the cost from the current node n to the goal node.

The classical A* algorithm involves the following steps:
\begin{enumerate}
\item Mark the starting node \textit{s} as “open” and calculate its cost \textit{f(s)};
\item Select the node \textit{n} from the open list that has the smallest value of \textit{f(n)};
\item If node \textit{n} is the goal node, terminate the algorithm;
\item Otherwise, mark node \textit{n} as “closed” and obtain a list of the neighbour nodes 
{\textit{n$_\textit{0}$} … \textit{n$_\textit{i}$}};
\item Calculate the costs \textit{f(n$_\textit{i}$)} for each neighbour node;
\item Mark all neighbour nodes as “open” except for those already in the closed list;
\item Remark as “open” any neighbour nodes that are marked as “closed” and have a smaller value \textit{f(n$_\textit{i}$)} than \textit{f(n)};
\item Return to the step 2.
\end{enumerate}

In this context, “open list” refers to a list of nodes that are yet to be evaluated, while “closed list” refers to a list of nodes that have already been evaluated. The cost function \textit{f(n)} calculates the distance between nodes using either Euclidean or Manhattan distance \cite{sharma_understanding_2020}. Based on the result of the cost function A* provides a path that is considered to be the best under current environmental conditions.

The classical approach takes into account only distance between nodes, while we are interested in multi-criteria assessment. So, we changed the assessment function \textit{f(n)} in Equation~(\ref{eq:eq_2}) to the multi-fitness function in Equation~(\ref{eq:eq_1}) for our experiment.

Using the same evaluation function i.e. Equation~(\ref{eq:eq_1}) for both path planning algorithms ensures a fair comparison of the two algorithms applied. 

\section{Experimental Setup}
\begin{figure}[h]
    \centering
    \includegraphics[width=5cm]{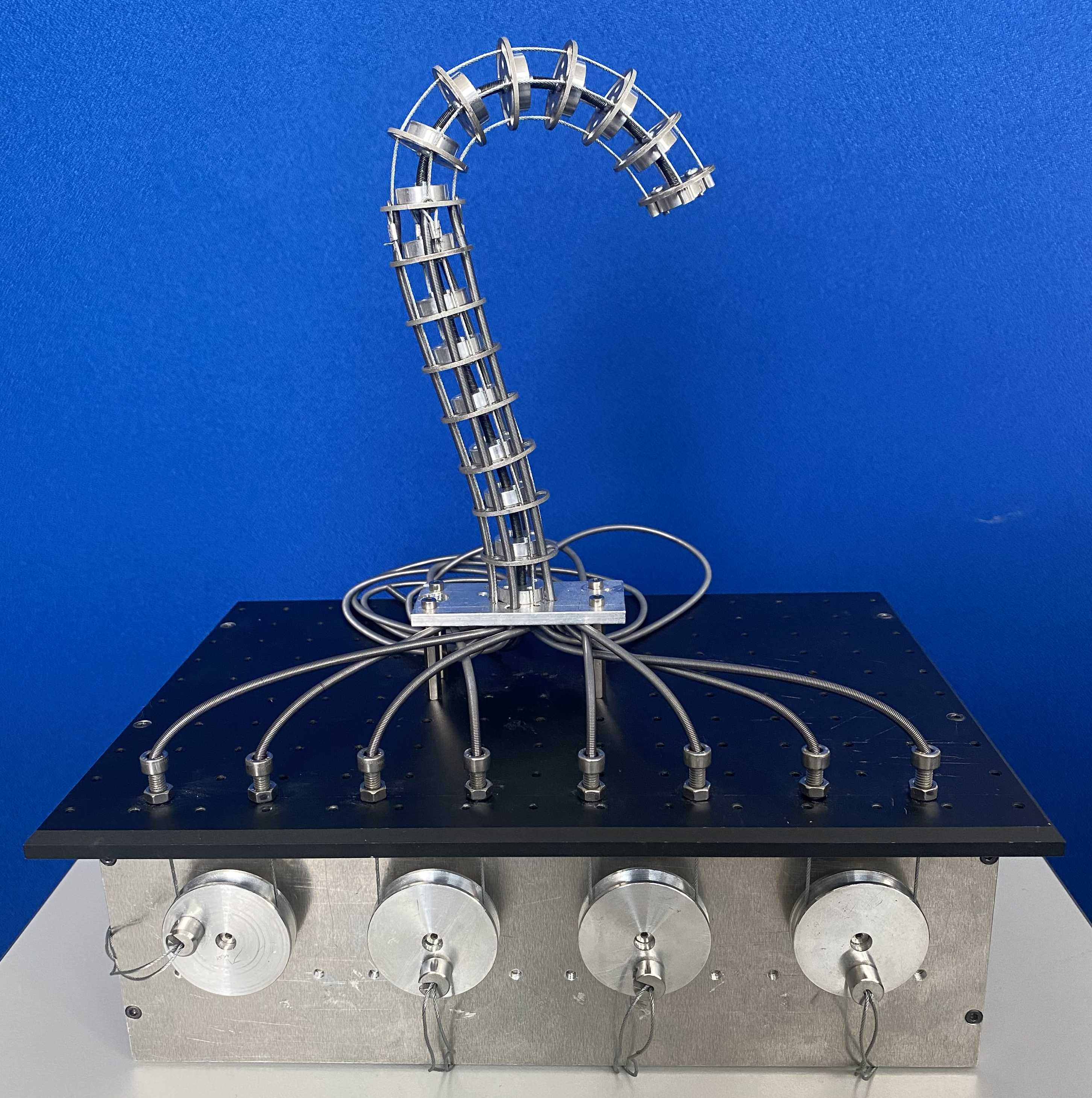}
    \caption{Self-built tendon-driven continuum robot prototype.}
    \label{fig:fig_1}
\end{figure}

For our experiments we use a self-built tendon-driven continuum robot prototype (see Figure~\ref{fig:fig_1}). Our prototype consists of 15 disks, one central flexible backbone tendon, and 8 side tendons. The disks are rigidly fixed on the central backbone tendon at an equal distance from each other. The prototype has a fixed base and two mobile sections. Side tendons pass through each disk along both sections (4 side tendons for each section), so that each section can be controlled independently \cite{shamilyan_meco_2022, shamilyan_access_2023}. The fixed base does not allow the robot itself to move, but the mobile sections can move in all possible directions. The lower section determines the starting point of the upper section, which covers a larger area, than the lower section does.

The robot arm of our prototype is actuated by 4 stepper motors (2 motors for each section). Using of stepper motors helps us to reduce the flexibility of the robot and make it more controllable. Each motor needs 800 steps per revolution which makes robot’s movements discrete. The stepper motors are equipped with absolute encoders that ensure the control of the number of steps taken by each motor. The stepper motors are connected to a microcontroller board that controls rotations of the stepper motors. 

Path planning calculations are executed on a laptop equipped with an Intel Core i5-10310U CPU @ 1.70GHz Processor and 8 GB RAM. The microcontroller board and the laptop communicate via the UART protocol, while the encoders and board utilize the SPI protocol.

\section{Experimental Evaluation}
The main idea of our experiment is to evaluate the paths generated by GA and A* under different criteria. We analyse the output of both algorithms to discover which one manages to get better results for multi-criteria optimization problems. To do this, we evaluated the time performance of the algorithms and the quality of the generated paths. The quality value is represented by the value of the fitness function. To conduct the experiment, we implemented a virtual environment to simulate the algorithms' results.

\subsection{Simulated Environment}
The environment of the prototype is discrete, due to the mechanical abilities of the stepper motors that control the prototype’s movements. Thus, the environment consists of a finite number of points that the robot can reach. 

\begin{figure}[t]
\centering
\includegraphics[width=0.8\columnwidth]{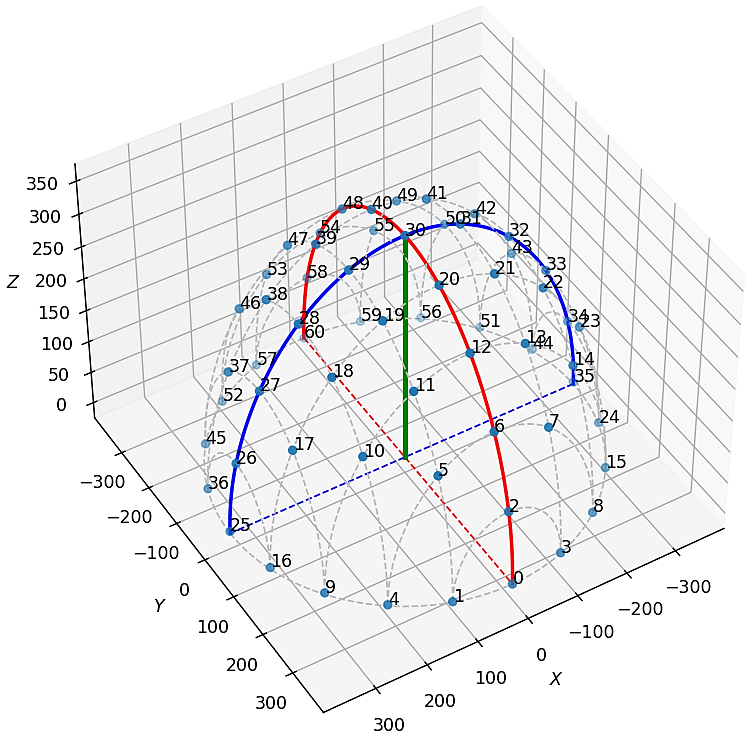}
\caption{Robot’s environment for one section.}
\label{fig:fig_2}
\end{figure}

Both sections are identical, and therefore, the environment for each section is also the same. Figure~\ref{fig:fig_2} shows the environment of one of the sections. The simulated environment created in Python and has been drawn using the matplotlib library. The local environment of the section resembles half of a sphere, due to the robot's physical peculiarities. The blue and red curved lines represent the routes along the X and Y axes, respectively. The straight green line indicates the position of one section of the robot. The section can move along grey dashed lines and stop at the blue nodes, each with its own unique id number. Each pair of adjacent nodes along the axes is separated by a distance of 70 motor steps. The smaller the distance between the nodes, the less discrete the map and the higher the map density. With a 70-step interval, the robot's environment includes 61 nodes for one section and a total of 3721 nodes (61 section nodes * 61 possible section positions) for all the possible positions of the upper section. Both, A* and GA, use the simulated environment to generate the path. To provide a small example of how paths are generated, let us take point 4 as the start node and point 6 as the goal node. In this case the shortest path will be 4-5-6, but there are also many other options, for example 4-10-11-12-6, 4-1-2-6 etc. 

\subsection{Alternative Paths}
\begin{figure}[b]
\centering
\includegraphics[trim={0 0 0 2.3cm}, width=0.9\columnwidth, clip]{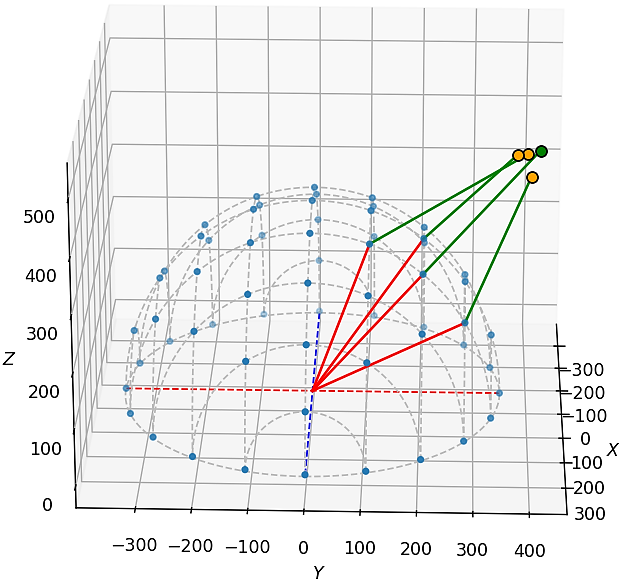}
\caption{Alternative goal points.}
\label{fig:fig_3}
\end{figure}

We implemented an algorithm to search for alternative goal points. For a given goal point the algorithm searches in the robot global environment for the three nearest neighbours. The usage of this algorithm increases the choice of possible endpoints and, as a consequence, increases the diversity of possible paths generated by path planning algorithms. The increased set of available paths gives more options to choose the best path for the robot, given the current state of the robot, of its environment and given criteria weights. The alternatives search algorithm provides the same output data for GA and A*. The distance between the initial and alternative points is calculated using the Euclidean distance. Figure~\ref{fig:fig_3} shows an example. Here the goal point is marked green and the nearest neighbours are marked orange. Red lines represent the position of the lower section of the robot, and green lines represent the position of the upper section. For a clearer image we omit to display the upper section´s environment. 

\subsection{Criteria Analysis}

The AHP algorithm calculates weights for each criterion based on their relative importance and uses these weights to evaluate the generated paths. We conducted multiple experiments for each of the possible combinations of the criteria given in Table \ref{tab:Table2}. Table \ref{tab:Table3} displays the criteria combinations that were used, with each line representing a single combination and each cell indicating the weight of the corresponding criterion. The green cells indicate the prioritised criteria, i.e. criteria which have more weight. The case in which no criteria were prioritised is not included, as the weight values are the same as in the case number 15. Prioritizing no or all criteria means they have equal relative importance levels which means weight calculation has no effect. Criteria weights from the Table \ref{tab:Table3} are used in Equation~(\ref{eq:eq_1}) to calculate the total fitness value of the generated path.

There are four different fitness functions for each of the criteria. The fitness value “distance” is calculated as the sum of Euclidean distances between each node in the generated path. “Motors' damage” fitness function calculates how many steps each motor will make if the robot follows the generated path. The “Mechanical damage” fitness function calculates how many times each segment between two nodes was used. In order to evaluate the “accuracy” the fitness function calculates the Euclidean distances between the planned/intended goal point and actually reached point. Quantitative characteristics for calculating the fitness functions of motor and mechanical damage criteria are stored in an external sqlite3 database. 

We also conducted an experiment to analyse how the generated paths vary depending on prioritised criteria. For this experiment we used just one path option without alternative goal points. The start and goal points for the lower section are 50 and 3, for the upper section 47 and 14. The start and goal points of both sections are randomly selected.

\begin{figure*}[t]
\centering
\includegraphics[width=\textwidth]{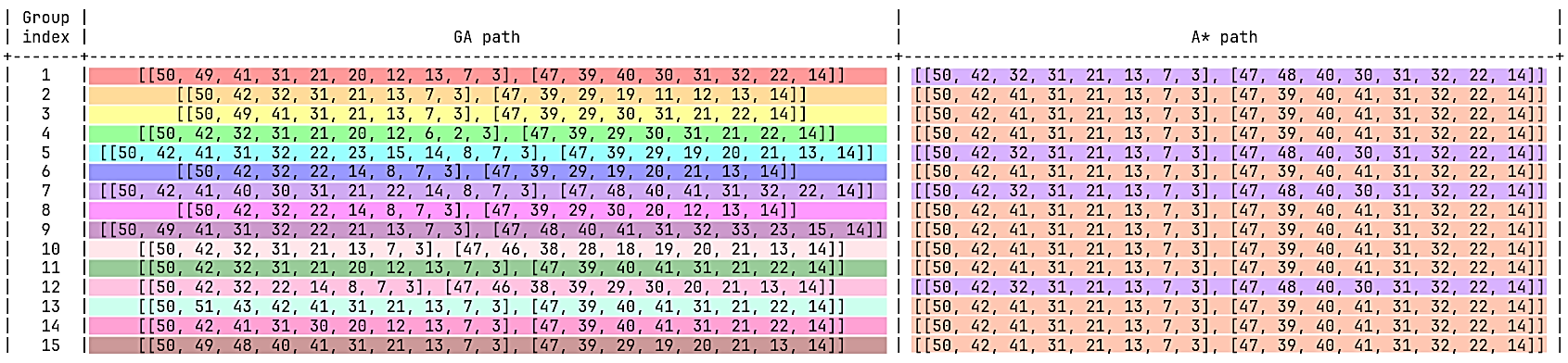}
\caption{Criteria analysis results. Each line shows the generated path for a particular criteria combination group (Group index). Different paths are highlighted with different colors. Column “GA path” shows final results of the GA. Column “A* path” shows final results of the A* algorithm. Comparing both columns clearly shows that GA provides a higher variety of paths.}
\label{fig:fig_4}
\end{figure*}

Figure~\ref{fig:fig_4} shows the results of the criteria analysis. The “group index” column refers to the corresponding criteria combination from the Table \ref{tab:Table3}. “GA path” and “A* path” columns contain paths generated by the two path-planning algorithms. Paths are represented as 2D arrays. The first part of each array refers to the path, generated for the lower section of the robot, and the second part – for the upper section. Different paths are marked in different colours.

The results show that GA is more sensitive to criteria variations than A*. For the given start and goal points, GA generated different paths for each criteria group, whereas A* generated only two different variants of paths.

\begin{table}[t]
  \centering
  \caption{Weights for different combinations of criteria (prioritized criteria are highlighted in light green)}
    \begin{tabularx}{\columnwidth}{|Y|Y|Y|Y|Y|}
    \hline
    \textbf{Group index} & \textbf{Distance} & \textbf{Motors damage} & \textbf{Mechanical damage} & \textbf{Accuracy}\\
    \hline
    \multicolumn{5}{|c|}{\textbf{\textit{Single criterion}}} \\
    \hline
    1 & \cellcolor{green}0.75 & 0.083 & 0.083 & 0.083 \\
    \hline
    2 & 0.083 & \cellcolor{green}0.75 & 0.083 & 0.083 \\
    \hline
    3 & 0.083 & 0.083 & \cellcolor{green}0.75 & 0.083 \\
    \hline
    4 & 0.083 & 0.083 & 0.083 & \cellcolor{green}0.75 \\
    \hline
    \multicolumn{5}{|c|}{\textbf{\textit{Two criteria}}}\\
    \hline
    5 & \cellcolor{green}0.45 & \cellcolor{green}0.45 & 0.05 & 0.05\\
    \hline
    6 &	\cellcolor{green}0.45 & 0.05 & \cellcolor{green}0.45 & 0.05\\
    \hline
    7 &	\cellcolor{green}0.45 & 0.05 & 0.05 & \cellcolor{green}0.45\\
    \hline
    8 &	0.05 & \cellcolor{green}0.45 & \cellcolor{green}0.45 & 0.05\\
    \hline
    9 &	0.05 & \cellcolor{green}0.45 & 0.05 & \cellcolor{green}0.45\\
    \hline
    10 & 0.05 &	0.05 & \cellcolor{green}0.45 & \cellcolor{green}0.45\\
    \hline
    \multicolumn{5}{|c|}{\textbf{\textit{Three criteria}}}\\
    \hline
    11 & \cellcolor{green}0.321 & \cellcolor{green}0.321 & \cellcolor{green}0.321 & 0.036\\
    \hline
    12 & \cellcolor{green}0.321 & \cellcolor{green}0.321 & 0.036 & \cellcolor{green}0.321\\
    \hline
    13 & \cellcolor{green}0.321 & 0.036 & \cellcolor{green}0.321 & \cellcolor{green}0.321\\
    \hline
    14 & 0.036 & \cellcolor{green}0.321 & \cellcolor{green}0.321 & \cellcolor{green}0.321\\
    \hline
    \multicolumn{5}{|c|}{\textbf{\textit{Four criteria}}}\\
    \hline
    15 & \cellcolor{green}1 & \cellcolor{green}1 & \cellcolor{green}1 & \cellcolor{green}1\\
    \hline
  \end{tabularx}
  \label{tab:Table3}
\end{table}

\subsection{Performance Analysis}

In this section we present results of the performance analysis of GA and A*. Both algorithms performed the same task under the same conditions and for the same input data. We divided the experiment into two parts: “single goal point” and “multiple goal point”. For the “single goal point” experiment, both algorithms generate paths using only the given start and goal points, without using alternative goal points. For the “multiple goal point” experiment, the alternative search algorithm is used. For both experiments all time measurements are shown as an average value of 100 algorithm runs. The axis “group index” refers to values from Table \ref{tab:Table3}.

\begin{figure}[b]
\centering
\includegraphics[width=\columnwidth]{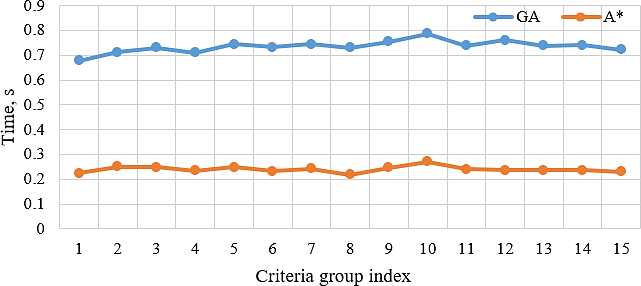}\\
(a)\\
\includegraphics[width=\columnwidth]{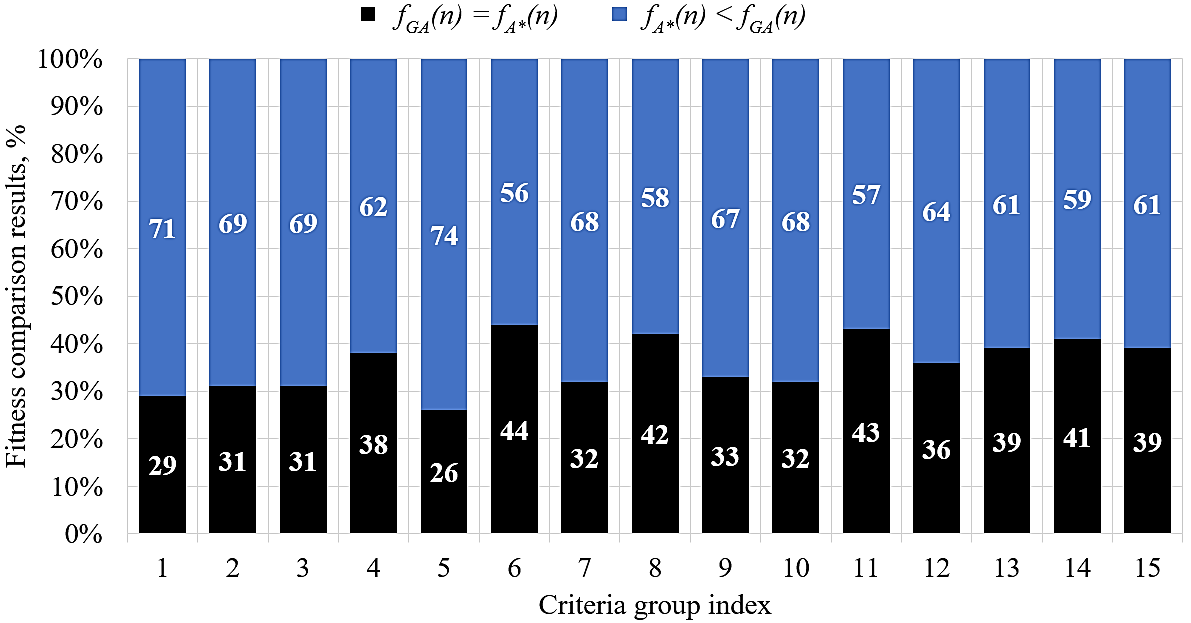}\\
(b)
\caption{“Single goal point” Experiment: a) Processing Time b) Relation of solutions with better fitness: A* better than GA (blue), both yielding the same results (black).}
\label{fig:fig_5}
\end{figure}

The results of the “single goal point” experiments are illustrated in Figure~\ref{fig:fig_5}. Figure~\ref{fig:fig_5}a shows the time measurements of both algorithms for each combination of criteria. A* algorithm takes twice less time than GA regardless of the criteria combination. Figure~\ref{fig:fig_5}b shows which algorithm achieved more often a superior fitness function value than the other, displayed are the percentages for both algorithms. From the results in Figure~\ref{fig:fig_5}b we see that even though GA creates more path variability for the different criteria groups than A*, it does not show efficiency in terms of time and fitness performance. For all 15 criteria groups A* generates better paths than GA in almost 60\% of the cases (blue bars). In other cases GA and A* have same fitness values (black bars) means both algorithms generated the same paths.

\begin{figure}[b]
\centering
\includegraphics[width=\columnwidth]{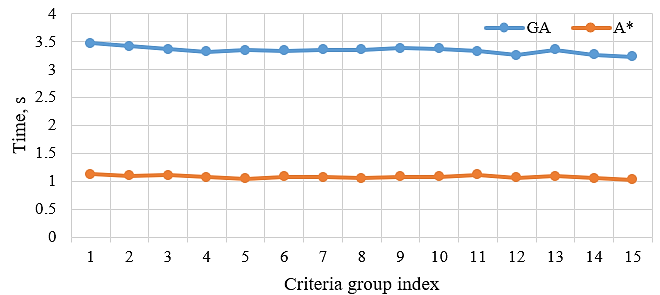}\\
(a)\\
\includegraphics[width=\columnwidth]{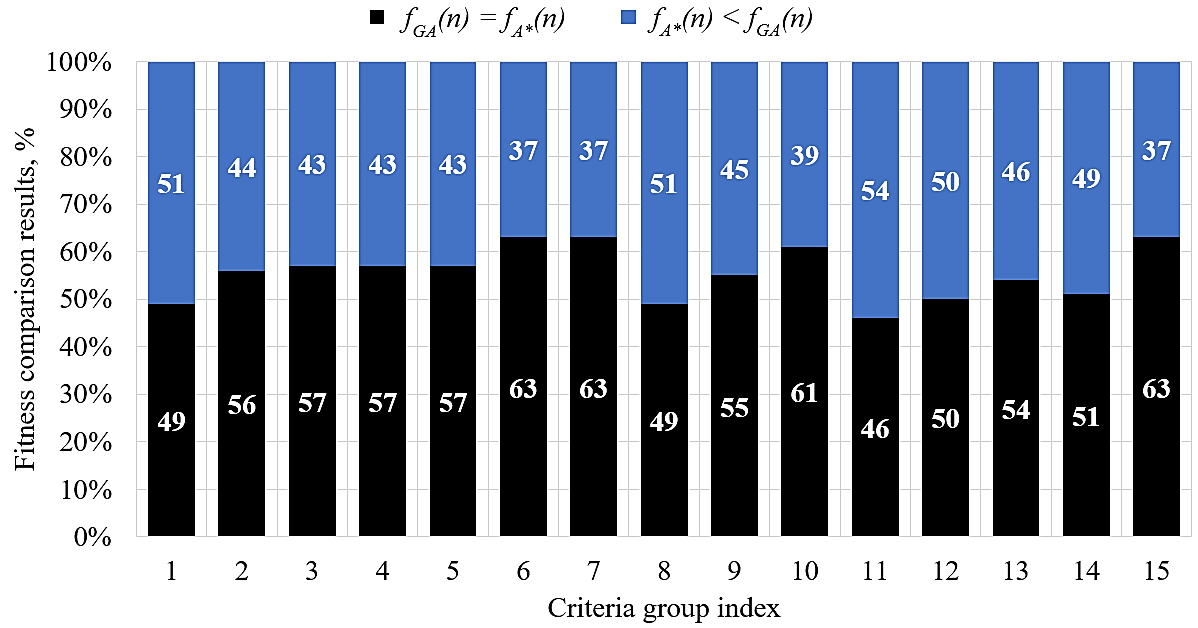}\\
(b)\\
\caption{“Multiple goal point” experiment: a) Processing Time, b) Relation of solutions with better fitness: A* better than GA (blue), both yielding the same results (black).}
\label{fig:fig_6}
\end{figure}

The “multiple goal point” experiment involves the use of the alternative search algorithm. The output of this algorithm are three additional goal points representing the closest neighbours of the initial goal point. Using alternative search, the path planning algorithms generate paths to 4 different goal points (1 initial + 3 alternatives) and then choose the one with the lowest fitness value. On the one hand this approach leads to an increased processing time, but on the other hand it provides a large variety of choices. The results of the “multiple goal point” experiments are illustrated in Figure~\ref{fig:fig_6}.

Figure~\ref{fig:fig_6}a shows a comparison of the execution times for both algorithms. As expected, the execution times of both algorithms regardless of criteria is proportionally increased by 4 times in accordance with the increase in the number of goal points. Figure~\ref{fig:fig_6}b shows the comparison of the fitness values. As well as in Figure~\ref{fig:fig_5}b, GA does not show better fitness results than A*, but the number of cases where GA and A* generated paths with same fitness value is increased almost 1.5 times (black bars). This can be explained by the fact that if there are more paths to be generated, the likelihood of generating identical paths increases.

The performance advantage of the A* over GA can be explained by the difference in algorithms structure. A* calculates the fitness function during path generation, and GA first generates the entire path and then calculates its fitness.

\subsection{Comparison of improved and classical implementation of path-planning algorithms}

\begin{figure}[t!]
\centering
\includegraphics[width=8.5cm]{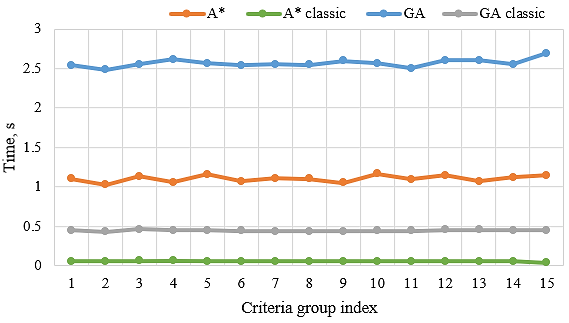}\\
\caption{Comparison of processing time between A* (orange), classical A* (green), GA (blue) and classical GA (gray).}
\label{fig:fig_7}
\end{figure}

\begin{figure*}[t]
\centering
\includegraphics[width=\textwidth]{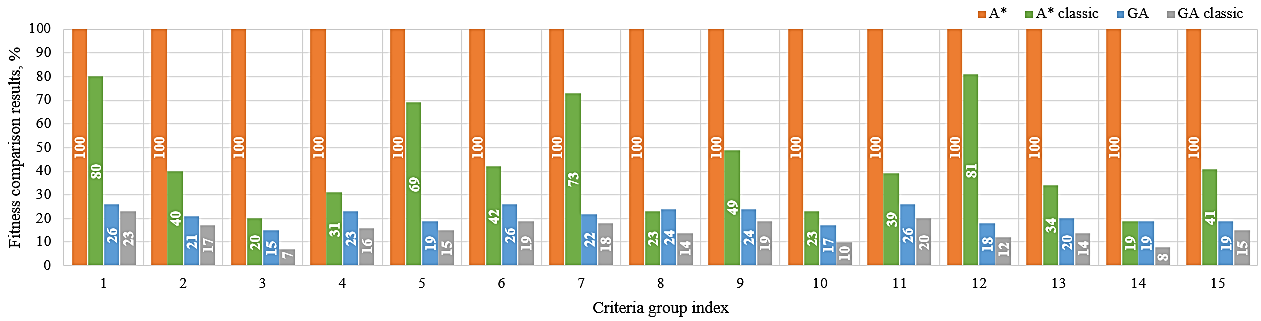}
\caption{Relation of solutions with better fitness: A* (orange bars), classical A* (green bars), GA (blue bars) and classical GA (gray bars).}
\label{fig:fig_8}
\end{figure*}

To ensure that our improved versions of the  algorithms provide enhanced performance, we provide a time and fitness comparison with the classical version of GA and A* algorithm (see Figure~\ref{fig:fig_7} and Figure~\ref{fig:fig_8}). The main difference between classical and improved versions is that the classical algorithm generates the path based only on the calculation of the path distance. Even thought classical implementation of both algorithms work significantly faster (see Figure~\ref{fig:fig_7}), it does not show performance superiority (see Figure~\ref{fig:fig_8}) and does not reflect criteria priorities. Figure~\ref{fig:fig_8} shows how many percents of the paths generated by each algorithm achieved the best fitness value. To calculate these results we run each algorithm 100 times for each criteria group using the same goal and start points, in order to provide clear result data. For example, for the first criteria group A* generated the best path in 100\% of all cases, while classical A* yielded the same results as A* in 80\% of all cases, GA achieved this in 26\% of cases, while classical GA did so in 23\%. It is important to emphasise that neither the classical A* algorithm nor GA nor the classical GA demonstrated results, which were better than the improved A* algorithm. At best they generated the same path as A* did.

\section{Conclusion}

Recently, resilience became a very important feature, due to the growing demand for autonomous systems. Such systems should be self-aware and have the ability to adapt their behaviour \cite{shamilyan_access_2023}. In our research of continuum robots, path planning is considered as an essential task for the system. Therefore, the path should be generated according to the system and environment state. 

This paper discusses an experiment with improved versions of GA and A* algorithm. We took classical versions of both algorithms and modified them by adding decision-making algorithm AHP to allow for more criteria than the path length to be considered when assessing the quality of the path calculated. In addition we evaluated  the algorithms if they may use alternative goal points to better adapt to the criteria. Both modifications are aiming to increase the resilience properties of the system. The AHP algorithm generates weights to the criteria of choice, thus helps to generate the path that fits the most. The alternatives search gives the opportunity to generate multiple paths to create more variety of choice. The paper provides a comparison between classical and improved versions of GA and A* as well as a comparison of two improved algorithms. Results show that improved versions of algorithms show better performance than their classical versions, and in particular, the improved A* algorithm outperforms all others.

Our experiments clearly show that it is feasible to enable an autonomous robot to consider complex parameters when making decisions. Here we showcased this for example by taking mechanical damage of its motor and tendons into account. This led to calculating paths completely different from the shortest ones, but with less negative impact on the robot. 

In our future research we are planning to provide feedback about how the “system aging” can influence the decision-making to adapt path-planning results, i.e. with increased percentage of mechanical and motors damage, the robot should follow the path different from the shortest one, based on the current state of the robot in order to minimize the load and increase its remaining useful life. Furthermore, we intend to investigate the potential scalability of the developed algorithms and apply them to a larger robot with an increased number of sections.

%%%%%%%%%%%%%%%%%%%%%%%%%%%%%%%%%%%%%%%%%%%%%%%%%%%%%%%%%%%%%%%%%%%%%%%%

\bibliography{mybibfile}

\begin{thebibliography}{21}
\providecommand{\natexlab}[1]{#1}
\providecommand{\url}[1]{\texttt{#1}}
\expandafter\ifx\csname urlstyle\endcsname\relax
  \providecommand{\doi}[1]{doi: #1}\else
  \providecommand{\doi}{doi: \begingroup \urlstyle{rm}\Url}\fi

\bibitem[Barnouti et~al.(2016)Barnouti, Al-Dabbagh, and
  Sahib~Naser]{barnouti_pathfinding_2016}
N.~H. Barnouti, S.~S.~M. Al-Dabbagh, and M.~A. Sahib~Naser.
\newblock Pathfinding in {Strategy} {Games} and {Maze} {Solving} {Using} {A}*
  {Search} {Algorithm}.
\newblock \emph{Journal of Computer and Communications}, 04\penalty0
  (11):\penalty0 15--25, 2016.
\newblock \doi{10.4236/jcc.2016.411002}.

\bibitem[Bhoskar et~al.(2015)Bhoskar, Kulkarni, Kulkarni, Patekar, Kakandikar,
  and Nandedkar]{bhoskar_genetic_2015}
M.~T. Bhoskar, M.~O.~K. Kulkarni, M.~N.~K. Kulkarni, M.~S.~L. Patekar,
  G.~Kakandikar, and V.~Nandedkar.
\newblock Genetic algorithm and its applications to mechanical engineering: A
  review.
\newblock \emph{Materials Today: Proceedings}, 2\penalty0 (4):\penalty0
  2624--2630, 2015.
\newblock \doi{https://doi.org/10.1016/j.matpr.2015.07.219}.

\bibitem[Chikhaoui et~al.(2019)Chikhaoui, Lilge, Kleinschmidt, and
  Burgner-Kahrs]{intro_6}
M.~T. Chikhaoui, S.~Lilge, S.~Kleinschmidt, and J.~Burgner-Kahrs.
\newblock Comparison of modeling approaches for a tendon actuated continuum
  robot with three extensible segments.
\newblock \emph{IEEE Robotics and Automation Letters}, 4\penalty0 (2):\penalty0
  989--996, 2019.
\newblock \doi{10.1109/LRA.2019.2893610}.

\bibitem[Cosserat and Cosserat(1909)]{cosserat}
E.~Cosserat and F.~Cosserat.
\newblock Th{\'e}orie des corps d{\'e}formables.
\newblock \emph{Nature}, 81\penalty0 (2072):\penalty0 67--67, Jul 1909.
\newblock ISSN 1476-4687.
\newblock \doi{10.1038/081067a0}.

\bibitem[Ghaheri et~al.(2015)Ghaheri, Shoar, Naderan, and
  Hoseini]{ghaheri_applications_2015}
A.~Ghaheri, S.~Shoar, M.~Naderan, and S.~S. Hoseini.
\newblock The {Applications} of {Genetic} {Algorithms} in {Medicine}.
\newblock \emph{Oman Medical Journal}, 30\penalty0 (6):\penalty0 406--416, Nov.
  2015.
\newblock \doi{10.5001/omj.2015.82}.

\bibitem[Guruji et~al.(2016)Guruji, Agarwal, and
  Parsediya]{guruji_time-efficient_2016}
A.~K. Guruji, H.~Agarwal, and D.~K. Parsediya.
\newblock Time-efficient {A}* {Algorithm} for {Robot} {Path} {Planning}.
\newblock \emph{Procedia Technology}, 23:\penalty0 144--149, Jan. 2016.
\newblock \doi{10.1016/j.protcy.2016.03.010}.

\bibitem[Hart et~al.(1968)Hart, Nilsson, and Raphael]{hart_formal_1968}
P.~E. Hart, N.~J. Nilsson, and B.~Raphael.
\newblock A {Formal} {Basis} for the {Heuristic} {Determination} of {Minimum}
  {Cost} {Paths}.
\newblock \emph{IEEE Transactions on Systems Science and Cybernetics},
  4\penalty0 (2):\penalty0 100--107, July 1968.
\newblock \doi{10.1109/TSSC.1968.300136}.

\bibitem[Holland(1992)]{ga_basics}
J.~H. Holland.
\newblock {A}daptation in {N}atural and {A}rtificial {S}ystems.
\newblock
  \url{https://mitpress.mit.edu/9780262581110/adaptation-in-natural-and-artificial-systems/},
  1992.
\newblock [Accessed 08-01-2024].

\bibitem[Jianqin and Xiao(2022)]{jianqin_research_2022}
L.~Jianqin and G.~Xiao.
\newblock Research on improved {A}-star algorithm for global path planning of
  unmanned logistics vehicles.
\newblock In \emph{2022 14th {International} {Conference} on {Intelligent}
  {Human}-{Machine} {Systems} and {Cybernetics} ({IHMSC})}, pages 44--47, Aug.
  2022.
\newblock \doi{10.1109/IHMSC55436.2022.00019}.

\bibitem[Kim et~al.(2020)Kim, Suh, and Han]{kim_development_2020}
C.~Kim, J.~Suh, and J.-H. Han.
\newblock Development of a {Hybrid} {Path} {Planning} {Algorithm} and a
  {Bio}-{Inspired} {Control} for an {Omni}-{Wheel} {Mobile} {Robot}.
\newblock \emph{Sensors}, 20\penalty0 (15):\penalty0 4258, July 2020.
\newblock \doi{10.3390/s20154258}.

\bibitem[Luca(2020)]{luca_roulette_2020}
G.~D. Luca.
\newblock {R}oulette {S}election in {G}enetic {A}lgorithms.
\newblock
  \url{https://www.baeldung.com/cs/genetic-algorithms-roulette-selection},
  2020.
\newblock [Accessed 27-02-2024].

\bibitem[Neumann and Burgner-Kahrs(2016)]{intro_3}
M.~Neumann and J.~Burgner-Kahrs.
\newblock Considerations for follow-the-leader motion of extensible
  tendon-driven continuum robots.
\newblock In \emph{2016 IEEE International Conference on Robotics and
  Automation (ICRA)}, pages 917--923, 2016.
\newblock \doi{10.1109/ICRA.2016.7487223}.

\bibitem[Nguyen and Burgner-Kahrs(2015)]{intro_1}
T.-D. Nguyen and J.~Burgner-Kahrs.
\newblock A tendon-driven continuum robot with extensible sections.
\newblock In \emph{2015 IEEE/RSJ International Conference on Intelligent Robots
  and Systems (IROS)}, pages 2130--2135, 2015.
\newblock \doi{10.1109/IROS.2015.7353661}.

\bibitem[Renda et~al.(2018)Renda, Boyer, Dias, and Seneviratne]{intro_2}
F.~Renda, F.~Boyer, J.~Dias, and L.~Seneviratne.
\newblock Discrete cosserat approach for multisection soft manipulator
  dynamics.
\newblock \emph{IEEE Transactions on Robotics}, 34\penalty0 (6):\penalty0
  1518--1533, 2018.
\newblock \doi{10.1109/TRO.2018.2868815}.

\bibitem[Rucker and Webster~III(2011)]{intro_5}
D.~C. Rucker and R.~J. Webster~III.
\newblock Statics and dynamics of continuum robots with general tendon routing
  and external loading.
\newblock \emph{IEEE Transactions on Robotics}, 27\penalty0 (6):\penalty0
  1033--1044, 2011.
\newblock \doi{10.1109/TRO.2011.2160469}.

\bibitem[Rucker et~al.(2010)Rucker, Jones, and Webster~III]{intro_4}
D.~C. Rucker, B.~A. Jones, and R.~J. Webster~III.
\newblock A geometrically exact model for externally loaded concentric-tube
  continuum robots.
\newblock \emph{IEEE Transactions on Robotics}, 26\penalty0 (5):\penalty0
  769--780, 2010.
\newblock \doi{10.1109/TRO.2010.2062570}.

\bibitem[Saaty(1980)]{saaty_analytic_1980}
T.~L. Saaty.
\newblock \emph{The {Analytic} {Hierarchy} {Process}: {Planning}, {Priority}
  {Setting}, {Resource} {Allocation}}.
\newblock McGraw-Hill International Book Company, 1980.

\bibitem[Shamilyan et~al.(2022)Shamilyan, Kabin, Dyka, and
  Langendoerfer]{shamilyan_meco_2022}
O.~Shamilyan, I.~Kabin, Z.~Dyka, and P.~Langendoerfer.
\newblock Distributed {Artificial} {Intelligence} as a {Means} to {Achieve}
  {Self}-{X}-{Functions} for {Increasing} {Resilience}: the {First} {Steps}.
\newblock In \emph{2022 11th {Mediterranean} {Conference} on {Embedded}
  {Computing} ({MECO})}, pages 1--6, June 2022.
\newblock \doi{10.1109/MECO55406.2022.9797193}.

\bibitem[Shamilyan et~al.(2023)Shamilyan, Kabin, Dyka, Sudakov, Cherninskyi,
  Brzozowski, and Langendoerfer]{shamilyan_access_2023}
O.~Shamilyan, I.~Kabin, Z.~Dyka, O.~Sudakov, A.~Cherninskyi, M.~Brzozowski, and
  P.~Langendoerfer.
\newblock Intelligence and {Motion} {Models} of {Continuum} {Robots}: {An}
  {Overview}.
\newblock \emph{IEEE Access}, 11:\penalty0 60988--61003, 2023.
\newblock \doi{10.1109/ACCESS.2023.3286300}.

\bibitem[Sharma(2024)]{sharma_understanding_2020}
P.~Sharma.
\newblock {U}nderstanding {D}istance {M}etrics {U}sed in {M}achine {L}earning.
\newblock
  \url{https://www.analyticsvidhya.com/blog/2020/02/4-types-of-distance-metrics-in-machine-learning/},
  2024.
\newblock [Accessed 07-03-2024].

\bibitem[Zuo et~al.(2015)Zuo, Guo, Xu, and Fu]{zuo_hierarchical_2015}
L.~Zuo, Q.~Guo, X.~Xu, and H.~Fu.
\newblock A hierarchical path planning approach based on a⁎ and least-squares
  policy iteration for mobile robots.
\newblock \emph{Neurocomputing}, 170:\penalty0 257--266, 2015.
\newblock \doi{https://doi.org/10.1016/j.neucom.2014.09.092}.

\end{thebibliography}

\end{document}